\documentclass{article}

\usepackage[preprint]{neurips_2019}

\usepackage{amsmath,amsfonts,bm}

\def\eqref#1{equation~\ref{#1}}

\def\1{\bm{1}}

\DeclareMathAlphabet{\mathsfit}{\encodingdefault}{\sfdefault}{m}{sl}
\SetMathAlphabet{\mathsfit}{bold}{\encodingdefault}{\sfdefault}{bx}{n}

\usepackage[utf8]{inputenc} 
\usepackage[T1]{fontenc}    
\usepackage{hyperref}       
\usepackage{url}            
\usepackage{booktabs}       
\usepackage{amsfonts}       
\usepackage{nicefrac}       
\usepackage{microtype}      
\usepackage{xcolor}
\usepackage[bottom]{footmisc}
\usepackage{newunicodechar}
\usepackage[utf8]{inputenc}
\usepackage{rotating}

\usepackage{multirow, makecell}
\usepackage{threeparttable}
\usepackage{footnote}
\usepackage{graphicx}
\usepackage{subfigure}
\makesavenoteenv{tabular}
\makesavenoteenv{table}

\newcommand{\hide}[1]{} 
\newcommand{\para}[1]{\noindent\textbf{#1 }}

\usepackage{stackengine}
\def\yenrule{\rule{1.3ex}{.1ex}}
\def\textyen{\renewcommand\stacktype{L}\stackon[.4ex]{\stackon[.65ex]{Y}{\yenrule}}{\yenrule}}

\title{Alchemy: A Quantum Chemistry Dataset for Benchmarking AI Models}

\author{%
  Guangyong Chen$^1$,~ 
  Pengfei Chen$^2$, 
  Chang-Yu Hsieh$^1$, 
  Chee-Kong Lee$^1$, \\
  \textbf{Benben Liao}$^1$, 
  \textbf{Renjie Liao}$^{3, 5}$,
  \textbf{Weiwen Liu}$^2$,
  \textbf{Jiezhong Qiu}$^4$,
  \textbf{Qiming Sun}$^1$, \\
  \textbf{Jie Tang}$^4$,
  \textbf{Richard Zemel}$^{3, 5}$,
  \textbf{Shengyu Zhang}$^1$ \\
Tencent$^1$, 
Chinese University of Hong Kong$^2$, \\
University of Toronto$^3$, 
Tsinghua University$^4$, 
Vector Institute$^5$ \\
\texttt{\{gycchen, kimhsieh, cheekonglee, bliao, qssun, shengyzhang\}@tencent.com} \\
\texttt{\{pfchen, wwliu\}@cse.cuhk.edu.hk} \\
\texttt{\{rjliao, zemel\}@cs.toronto.edu} \\
\texttt{qiujz16@mails.tsinghua.edu.cn}, \texttt{jietang@tsinghua.edu.cn}
}

\begin{document}

\maketitle

\begin{abstract}
We introduce a new molecular dataset, named Alchemy, for developing machine learning models useful in chemistry and material science. As of June 20th 2019, the dataset comprises of 12 quantum mechanical properties of 119,487 organic molecules with up to 14 heavy atoms, sampled from the GDB MedChem database. The Alchemy dataset expands the volume and diversity of existing molecular datasets. Our extensive benchmarks of the state-of-the-art graph neural network models on Alchemy clearly manifest the usefulness of new data in validating and developing machine learning models for chemistry and material science. We further launch a contest to attract attentions from researchers in the related fields. More details can be found on the contest website \footnote{https://alchemy.tencent.com}. At the time of benchamrking experiment, we have generated 119,487 molecules in our Alchemy dataset. More molecular samples are generated since then. Hence, we provide a list of molecules used in the reported benchmarks.
\end{abstract}

\section{Introduction}

Recent advances in machine learning~(ML) techniques have proven immensely useful for a broad range of applications including natural language processing (NLP)~\citep{shen2017reasonet}, computer vision~\citep{he2016deep} and strategic plannings~\citep{silver2016mastering,silver2017mastering} etc.  These remarkably successful demonstrations have drawn high interests from the physical and biological science communities. 
For instance, efficient generations of novel molecular structures~\citep{jin2018junction} under multi-objective optimizations and better strategies in synthesis planning~\citep{segler2018benevai} and rectrosynthesis~\citep{segler2017benevai} will significantly accelerate drug discovery and novel material design~\citep{sanchez2018science,gomez2018automatic}. A common component underlying these complex AI systems is a computational engine that could predict or compute molecular properties at high precision and speed. A successful delivery of these AI systems depends critically on this predictor under the hood. For instance, we need to assess the quality of generated molecules by checking the properties to be optimized, and use these predictions in a feedback loop to fine tune the generative model. Recently, the development of a ML-based molecular predictor has reached state of the art accuracy~\citep{gilmer2017neural,liao2019lanczosnet} using only a fraction of the computational resources typically required for a similar quantum chemical calculation.

The importance of supervised information for ML development cannot be understated.  The ImageNet~\citep{imagenet_cvpr09}, a collection of more than 1.5 million labelled images distributed over $1,000$ classes, facilitates the development of new model such as ResNet~\citep{he2016deep} that surpasses human performance in image recognition. In another instance, the Stanford Question Answering Dataset (SQuAD)~\citep{rajpurkar2016squad}, a reading comprehension dataset consisting of $150,000$ questions and answers, is critical to the development of a powerful language representation model BERT~\citep{devlin2018bert}. Recognizing the importance of high-quality data for ML model developments,  the chemistry community has recently compiled a comprehensive collection of benchmarking datasets, the MoleculeNet~\citep{wu2018moleculenet}, including a variety of supervised learning tasks.  Despite this effort by the Pande group at Stanford, the amount of data in the MoleculeNet is inadequate in comparison to the typical size of ML training datasets. For instance, there are less than 150K molecular entries for training models to predict the quantum mechanical properties of small organic molecules. The biggest dataset QM9 within MoleculeNet is further restricted to curating molecules composed of Hydrogen~(H), Carbon~(C), Nitrogen~(N), Oxygen~(O) and Florine~(F). Therefore, 
the highly successful results mentioned earlier are not guaranteed to generalize beyond the scope of existing datasets. A better data variety, such as the presence of more atom types and larger molecular size, can help to more thoroughly investigate and improve some aspects of ML models such as generalibility, transferability and few-shot learning capability.
 
To the end of creating better ML models for molecular sciences, we decide to create a new molecular dataset for the training and investigative purposes. At the time we perform the benchmarks reported in this work, our dataset, named Alchemy, contains 12 quantum mechanical properties of 119,487 organic molecules with up to 14 heavy atoms~(C, N, O, F, S and Cl) from the GDB MedChem database~\citep{ruddigkeit2012enumeration}. More molecular data have been generated since then. The quantum mechanical properties are calculated using the Python-based Simulations of Chemistry Framework (PySCF)~\citep{sun2018pyscf} with details given in Sec.~\ref{sec:aichemy}.
As compared to the full GDB-17 database, the MedChem subset contains molecules that are screened as being more likely to be useful for medicinal chemistry based on functional group and complexity considerations. Therefore, the Alchemy dataset has a more sharpened focus on medicinal chemistry in comparison to other quantum-chemistry datasets.  We anticipate Alchemy dataset to facilitate evaluation, benchmarking and development of ML methods for applications in chemistry and materials science. In fact, we are hosting a molecular property prediction challenge based on Alchemy dataset, namely the Alchemy contest \footnote{https://alchemy.tencent.com}, to engage more people in the development of better ML models for molecular sciences.

\section{Related Datasets}
\label{sec:related}



There are only a few molecular datasets specifically built for benchmarking ML models for chemistry and material science applications.  Recently, a collection of various datasets have been compiled and collectively branded as the MoleculeNet.  Since experimental results often contain measurement noise, we report our contributions of new datasets for multi-task learning of quantum mechanical properties of organic molecules. Hence, only 
the subset of quantum mechanical databsets within the MoleculeNet is summarized below. 

\para{QM7/QM7b}  The QM7 dataset contains 7,165 molecules, a subset of GDB-13, with at most 7 C,N,O,S atoms. The learning task is to predict a single electronic property, the atomization energy.  All the physical properties listed in the dataset were calculated with the ab-initio density function theory at the level of PEB0/tier2 basis set. QM7b is an extension. 13 additional properties are computed at different levels of accuracy (ZINDO, SCS, PBE0, GW). The data is further expanded to 7,211 molecules.

\para{QM8} This dataset provides electronic properties of 21,786 molecules comprising up to 8 C,N,O,F atoms, a subset of GDB-17. Not only ground-state electronic properties but also four excited-state properties at different levels of accuracy (TDDFT using PBE0 and CAM-B3LYP and CC2) are computed for multi-task learning. 

\para{QM9} This dataset contains geometric, energetic, electronic and thermodynamics (totalling 13) properties for 133,885 organic molecule comprising up to 9 non-hydrogen atom within the GDB-17 database. All physical properties are computed at the accuracy of B3LYP/6-31G(2df,p) based DFT. 

\begin{table}\label{tab:data}
\caption{Dataset Details: number of molecules and tasks }\vspace{2mm}
\centering
\small
\begin{threeparttable}
\begin{tabular}{c|ccccc}\toprule
\textbf{Dataset} & \textbf{Data Type}             & \textbf{\#Tasks} & \textbf{\#Molecules} & \textbf{Rec-Split} & \textbf{Heavy Atoms} \\ \midrule
QM7     & SMILES, 3D coordinates & 1       & 7,165       & Stratified & $\leq$7\\
QM7b    & 3D coordinates        & 14      & 7,211       & Random     & $\leq$7\\
QM8     & SMILES, 3D coordinates & 12      & 21,786      & Random     & $\leq$8\\
QM9     & SMILES, 3D coordinates & 12      & 133,885     & Random     & $\leq$9\\ \midrule
\textbf{Alchemy} & SMILES, 3D coordinates & 12      & \textbf{119,487}     & Stratified/Size  & \textbf{9-14}  \\\bottomrule
\end{tabular}
\end{threeparttable}
\small
\end{table}

\section{Literature Survey about Molecular Machine Learning}
\label{sec:survey}
\subsection{Molecular Representations: SMILES and Fingerprint}

Intuitively, a molecule is a stable 3D configurations of atoms connected by chemical bonds, as described in the Alchemy dataset. The essential features of a molecule can be concisely encoded as a molecular graph with each vertex representing an atom and each edge representing a bond.  As molecules come in different sizes, the associated graphs also vary in terms of number of nodes, edges and structures, which makes the definition of basic operations (such as convolution) rather elusive. For a long time,  most ML models are not good at handling this kind of graph-structured data. Two common approaches to overcome the challenge is to convert the molecular graphs into the simplified molecular-input line-entry system (SMILES) \citep{Weininger:1988:SCL:50989.50994} or the fingerprint  representation.

The Alchemy dataset also contains the SMILES string, which is a clever scheme to encode a 2D graph in terms of a 1D text string.  In this way, a molecule can be viewed as a sentence and some advanced NLP techniques such as recurrent neural network (RNN) and long short-term memory (LSTM) model can be applied to process molecules of various sizes. However, SMILES string is not designed for ML purposes and several issues arise when it is used in a ML context. First, that generation of new molecules by outputting a new SMILES string character by character often ends up with invalid SMILES strings which cannot be converted back into molecules \citep{jin2018junction}. Secondly, multiple valid and markedly different SMILES strings could correspond to a same molecule. Even though one can use a canonicalization to systematically pick one SMILES string as a reference for each molecule, there are multiple algorithms to do this which could pick different SMILES string as the reference. Hence, a direct comparison of ``distance" based on the SMILES string is not a good measure to quantify similarities between molecules.

On the other hand, molecular fingerprint aims to record the presence or absence of some particular substructures, which are chosen based on expert knowledge for some special tasks. Thus, we can represent a molecule in terms of a sequence of binary bits with fixed length. While the fixed-length fingerprint is amenable to most ML models, there are other challenges. The fingerprint representation necessarily neglects certain molecular substructure. To distinguish as many molecular substructures as possible in a fingerprint, one might be forced to use an extremely long binary sequence, which disturbs the training of learning methods because of the introduction of too many redundant features.
\subsection{Prediction of Molecular Properties by Deep Neural Networks}
 As mentioned earlier, a high-quality molecular predictor is a critical component underlying many complex AI systems used in chemical applications.  
We briefly survey the trend of evolving design of ML-based predictors in the field of drug discovery, since the Alchemy dataset carries a focus on medicinal chemistry. 
Deep Neural Networks (DNN) have attracted significant attention since the Merck Molecular Activity Challenge hosted in 2012. Shortly after the competition, performances of DNNs have been investigated thoroughly and have been shown to deliver superior predictions than standard random forest on a set of large diverse quantitative structure–activity relationships (QSAR) datasets that are taken from Merck’s drug discovery effort~\citep{Ma2015}. In another systematic benchmarking study, Ramsundar \textit{et al}. ~\citep{Ramsundar2017} further demonstrates that DNNs outperform most conventional ML methods such as SVM, RF, Naive Bayes or k-nearest neighbor on seven datasets from ChemBL, one of the largest manually curated chemical databases of bioactive molecules with drug-like properties. Nevertheless, for model developments, it is more common to resort to QM-series datasets because of experimental noise contained in the real-world datasets.

\subsection{Graph Neural Networks}
Recently, novel techniques such as Graph Neural Networks (GNNs) \citep{gori2005new,scarselli2009graph} could automatically extract meaningful features from the molecular graphs without resorting to the more traditional approach of manually designing descriptors such as fingerprints.  Further benefits of a graph neural network approach goes beyond just an end-to-end learning of a data-driven molecular representation.  One can incorporate more microscopic details, such as  atomic pairwise distances, into the model. In this way, one could build a more robust and accurate molecular predictor \citep{li2016gated,kipf2017semi,gilmer2017neural,velivckovic2018graph,liao2019lanczosnet}.

GNNs broadly follow a recursive neighborhood aggregation procedure~\citep{xu2018representation,xu2019powerful}, or alternatively called message passing~\citep{gilmer2017neural}. During the forward propagation, the node state is updated recursively by aggregating and transforming hidden states of its neighboring nodes. For example, a hidden state is updated during the forward propagation stage of the Graph Convolutional Networks (GCN)~\citep{kipf2017semi} via, 
\begin{equation}
\label{Eg_gcn}
h_v^{(l+1)}=\sigma\left(\sum_{w\in \mathcal{N}_v}\dfrac{1}{c_{vw}}W_{1}^{(l)}h_w^{(l)} + W_{0}^{(l)}h_v^{(l)}\right),
\end{equation}
where $h_v^{(l)}$ denotes the hidden state of node $v$ at $l$-th layer, $W^{(l)}$ denotes trainable weight, $\sigma$ denotes an activation function such as ReLU, $c_{vw}$ is a normalization constant such as $c_{vw}=\sqrt{deg(v)deg(u)}$ and $deg(v)$ is the degree of node $v$.

Within the recursive aggregation framework, many efforts have been made to improve the expressive power of GNNs. For instance, 
Graph Attention Networks (GAT)~\citep{velivckovic2018graph} introduces an attention mechanism, which enables selective learning to identify important functional groups in a molecule. Relational GCN (RGCN)~\citep{schlichtkrull2018modeling} models multi-relational graph by learning a different weight matrix for each edge type. Gated Graph Neural Networks (GGNN)~\citep{li2016gated} treats hidden states of each node across layers as a sequence and update the states using Gated Recurrent Unit (GRU)~\citep{cho2014properties}. Message Passing Neural Network (MPNN)~\citep{gilmer2017neural} presents a unifying architecture that ties many variants of GNNs together. Moreover, it introduces an edge network which takes feature vectors of edges as input and gives the weights as output.  Lanczos Networks (LanczosNet)~\citep{liao2019lanczosnet} takes into account of multi-scale connections. It uses the Lanczos algorithm~\citep{lanczos1950iteration} to construct a low rank approximation of the graph Laplacian, enabling efficient exploitation of multi-scale information in graphs. Graph Isomorphism Network (GIN)~\citep{xu2019powerful} is a theoretical framework for analyzing the expressive power of GNNs and develops a simple and expressive architecture. When applying these models to molecules, a readout function~\citep{vinyals2015order,ying2018hierarchical} is implemented at the output layer of GNNs to obtain graph-level representations. Currently, there is an urgent need for a new dataset for quantum property regressions because the MPNN and its variants are highly optimized architectures for QM9 dataset. All aforementioned models have been implemented and benchmarked on the newly proposed dataset, Alchemy, in this work.


\section{The Alchemy Dataset}
\label{sec:aichemy}

\begin{figure}[tbp]
	\centering
	\mbox{
	\subfigure[OpenBabel geometry]{
	    \includegraphics[width=.2\columnwidth]{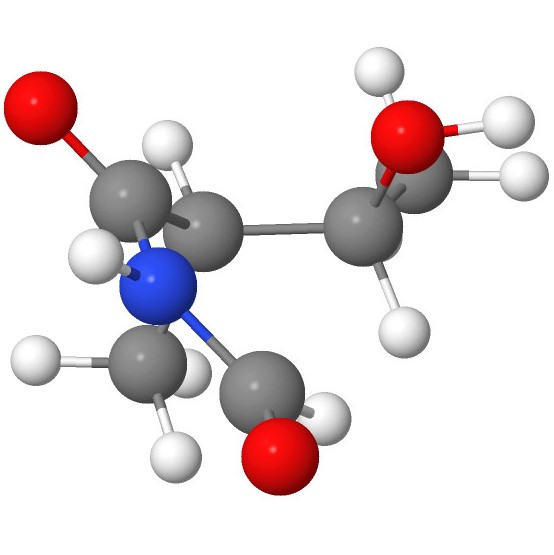}
	    \label{fig:openbabel}
	}
	\subfigure[HF/STO-3G geometry]{
	    \includegraphics[width=.2\columnwidth]{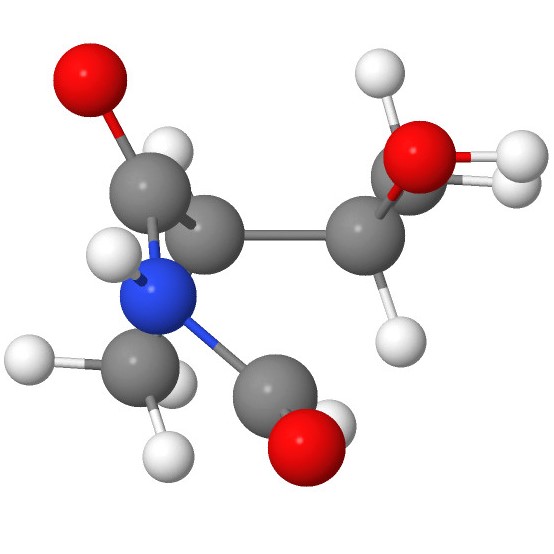}
	    \label{fig:sto3g}
	}
	\subfigure[DFT geometry]{
	    \includegraphics[width=.2\columnwidth]{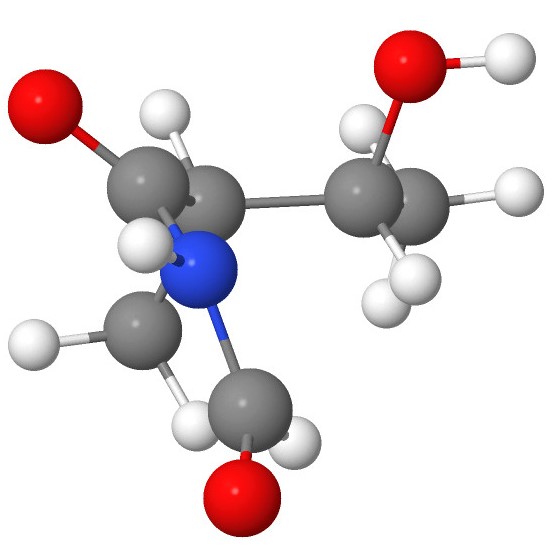}
	    \label{fig:dft}
	}
	}\small
	\caption{An example for molecule \textsf{CC(O)C(C)C(=O)NC=O}, a sample from Alchemy dataset.}
	\label{fig:example}\small
\end{figure}

\begin{figure}[tbp]
	\centering\small
    \includegraphics[width=1.\textwidth]{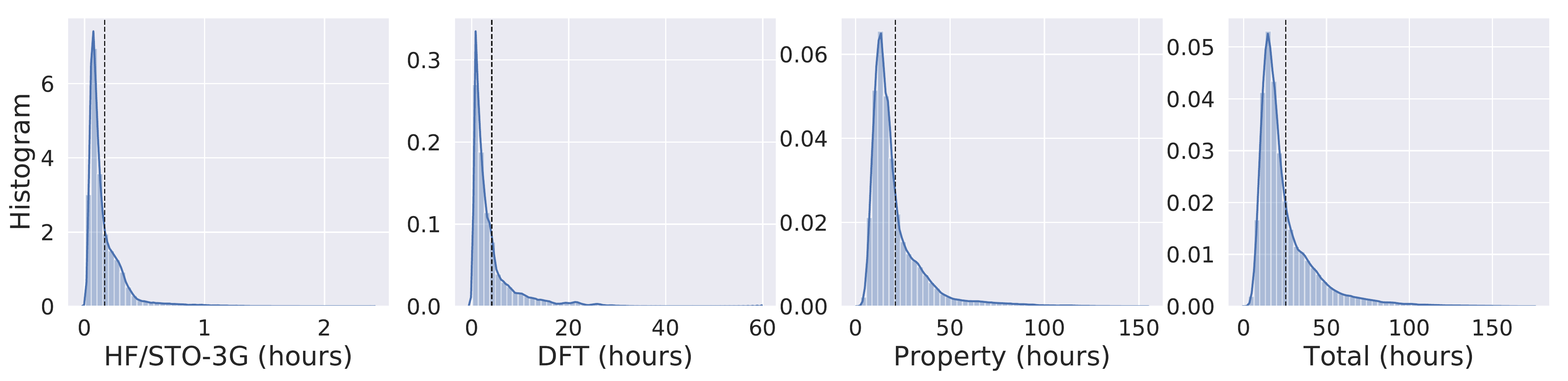}\small
	\caption{Histogram of Alchemy's running time. Each step takes 0.17/4.13/21.11 hours on average, respectively. The average total running time for processing a molecule is 25.41 hours.}
	\label{fig:time}\small
\end{figure}

\subsection{Workflow for the Data Generation}
The molecular properties were calculated using PySCF's implementation of the DFT Kohn-Sham method at the B3LYP level with the basis set 6-31G(2df,p). The quantum chemistry model B3LYP/6-31G(2df,p) was validated in an early work \citep{ramakrishnan2014quantum} for small organic molecules. Building a complementary dataset along
the line of QM-series introduced in Sec.~\ref{sec:related}, we computed the following categories of molecular properties: ground state equilibrium geometry, ground state electronic properties, and ground state thermochemical properties. The detail is summarized in Table \ref{tab:dft}. 

\hide{ 
\begin{table}
\centering
\caption{Calculated properties. Last two entries are not present in the QM9 dataset}
\label{tab:dft}
\small
\begin{tabular}[tp]{c|c|c|c}
\hline
\textbf{Property} & \textbf{Unit} & \textbf{Mean Relative Error} & \textbf{Meam Absolute Error}\\
\hline
Rotational constants & GHz & - & -\\
Dipole moment~(mu) & Debye & 0.1928 & 0.6401\\
Polarizability~(alpha) & $a_0^3$ & 0.0104 & 0.7610\\
HOMO & $E_\text{h}$ & 0.0244 & 0.0060\\
LUMO & $E_\text{h}$ & 0.2114 & 0.0061\\
gap &$E_\text{h}$ & 0.0275 & 0.0073\\
$\langle R^2\rangle$~(R2) & $a_0^2$ & 0.0404 & 52.3778\\
Zero point energy~(zpve) & $E_\text{h}$ & 0.0021 & 0.0003\\
Internal energy~(U0) & $E_\text{h}$ & 0.0006 & 0.2486 \\
Internal energy at 298.15 K~(U) & $E_\text{h}$ & 0.0006 & 0.2486\\
Enthalpy at 298.15 K~(H) & $E_\text{h}$ & 0.0006 & 0.2486\\
Free energy at 298.15 K~(G) & $E_\text{h}$ & 0.0006 & 0.2486\\
Heat capacity at 298.15 K~(Cv) & $E_\text{h}$K$^{-1}$ & 0.0038 & 3.8$\times 10^{-7}$\\
Dipole moment vector &- &- & - \\
3$\times$3 Polarizability tensor & -& - & -\\
\hline
\end{tabular}
\small
\end{table}
} 

In the remaining sections, we clarify a few technical aspects of our workflow for readers interested in quantum chemistry. The equilibrium geometry was optimized in three passes, as shown in Figure~\ref{fig:example}. We first used OpenBabel~\citep{O'Boyle2011} to parse SMILES string and built the Cartesian coordinates with MMFF94 force field optimization. The second pass employed the HF/STO-3G theory (incorporate quantum mechanical effects) to generate a preliminary geometry. In the final pass of geometry relaxation, we used the B3LYP/6-31G(2df,p) model with the density fitting approximation for electron repulsion integrals. The auxiliary basis cc-pVDZ-jkfit~\citep{Weigend2002} is employed in density fitting to build the Coulomb matrix and the HF exchange matrix. Alchemy's running time statistics can be found in  Figure~\ref{fig:time}, which shows that simulating quantum properties of molecules with 10 heavy atoms is quite time-consuming.

When computing the ground state electronic properties and ground state thermochemical properties, the density fitting technique was disabled in our dataset generation
program. In addition to the molecular properties proposed by the original QM9 dataset, we can also report the dipole moment vector, the 3$\times$3 polarizability
tensor and the atomic charges obtained by the meta-Lowdin population analysis~\citep{Sun2014}. The advantage of meta-lowdin population analysis, like other
advanced population analysis prescriptions~\citep{Knizia2013}, is that the obtained atomic charges possess good transferability in different molecular
environments.

In total, it takes about 3,000,000 CPU hours to generate all 119,487 molecules with heavy-atom count between 9 and 14 in this Alchemy dataset.

\begin{table}
\centering
\caption{Calculated properties by Alchemy and comparison with QM9.}
\label{tab:dft}
\small
\begin{threeparttable}
\begin{tabular}[tp]{c|c|c|c}
\toprule
\textbf{Property} & \textbf{Unit} & \textbf{Mean Relative Error} & \textbf{Meam Absolute Error}\\
\midrule
Dipole moment~(mu) & Debye & 0.1649 & 0.4741\\
Polarizability~(alpha) & $a_0^3$ & 0.0087 & 0.7180\\
HOMO & $E_\text{h}$ & 0.0210 & 0.0051 \\
LUMO & $E_\text{h}$ & 0.1869 & 0.0051\\
gap &$E_\text{h}$ &  0.0208 & 0.0058\\
$\langle R^2\rangle$~(R2) & $a_0^2$ & 0.0504 & 79.8315\\
Zero point energy~(zpve) & $E_\text{h}$ & 0.0017 & 0.0003\\
Internal energy~(U0) & $E_\text{h}$ & 0.0006 & 0.2643 \\
Internal energy at 298.15 K~(U) & $E_\text{h}$ & 0.0006 & 0.2643\\
Enthalpy at 298.15 K~(H) & $E_\text{h}$ & 0.0006 & 0.2643\\
Free energy at 298.15 K~(G) & $E_\text{h}$ & 0.0006 & 0.2644\\
Heat capacity at 298.15 K~(Cv) & $E_\text{h}$K$^{-1}$ & 0.0072 & 4.02$\times 10^{-7}$\\
\bottomrule
\end{tabular}
\end{threeparttable}
\small
\end{table}

\subsection{Comparison to the QM9 dataset}
To validate that we have successfully built a high-quality dataset, we first pick a set of 21,310 molecules from the Alchemy data that do not contain any S or Cl atoms. We then retrieve the same set of molecules from QM9 dataset.  We compare the molecular properties of these two datasets, and the results are summarized in Table~\ref{tab:dft}.
 We draw attention to the third and fourth columns of Table~\ref{tab:dft}.  The mean relative error and mean absolute error refers to the disagreement between our results and the original QM9. Our calculations agree well with the original QM9 given the fact that different definitions of B3LYP functionals were used in these calculations. Nevertheless, we notice a few disparities. The relative error on LUMO energy is less a concern as we get a good agreement on the HOMO-LUMO energy gap. It is well-known that the two definitions of B3LYP functionals tend to systematically shift orbital energies in these calculations, and the energy gap should be a more appropriate comparison criterion. However, inconsistency on dipole moment constitutes a more serious indications of disagreements. We unambiguously trace this inconsistency back to the complexity of searching the optimal molecular conformation among enormous molecular geometry local minimum. The geometry optimization could be stuck at the sub-optimal molecular conformation in either our dataset or QM9.

\subsection{File Format}
The data file format of QM9 is not the standard format to save molecular properties in the cheminformatics community. More precisely, QM9 does not provide bond information, i.e. which pairs of atoms are connected by chemical bonds in molecules. There should be no problem when one uses OpenBabel to read in QM9 data (which was originally processed with OpenBabel) and re-create a molecular graph with 3D atomic coordinates. It may be problematic if one uses RDKit to process the QM9 dataset since QM9 and RDKit handle the sequence of atoms of SMILES strings in different orders.  Therefore, errors could arise when RDKit tries to re-create a molecular graph with the 3D atomic coordinates given in an order set by OpenBabel. Therefore, we save the molecular data in the SD file format, which provides the bond information and avoids the ambiguity issue. Since most GNN models do not use 3D atomic coordinates, many earlier works did not actually use this piece of information. However, we caution that the situation could change in the near future as newer models, such as MPNN, takes in atomic pairwise distances as edge attributes. Therefore, one either has to use OpenBabel to process QM9 dataset or they could use the Alchemy dataset without having to worry about the file format.

\section{The Benchmarking Results}
\label{sec:task}

We conduct two benchmarking experiments with the following baselines: Graph Convolutional Networks (GCN)~\citep{kipf2017semi},  Chebyshev networks (ChebyNet)~\citep{defferrard2016convolutional}, Graph Attention Network (GAT)~\citep{velivckovic2018graph}, Relational Graph Convolutional Networks (RGCN)~\citep{schlichtkrull2018modeling}, Gated Graph Neural Networks (GGNN)~\citep{li2016gated}, Lanczos Networks (LanczosNet)~\citep{liao2019lanczosnet}, Graph Isomorphism Network (GIN)~\citep{xu2019powerful}, and Message Passing Neural Networks (MPNN)~\citep{gilmer2017neural}.

\begin{table}[t]
\small
\centering
\begin{threeparttable}
\caption{Node and Edge Features in Alchemy benchmarking experiments}\label{tab:feat}
\begin{tabular}{ccc}
\toprule
\textbf{Type}                  & \textbf{Feature}                   & \textbf{Description}                                \\
\midrule
\multirow{7}{*}{Node} & Atom type                 & H, C, N, O, F, \textbf{S}, \textbf{Cl} (one-hot)             \\
                      & Atomic number             & Number of protons (integer)                \\
                      & Acceptor                  & Accepts electrons (binary)                 \\
                      & Donor                     & Donates electrons (binary)                 \\
                      & Aromatic                  & In an aromatic system (binary)             \\
                      & Hybridization             & sp, sp2, sp3 (one-hot or null)             \\
                      & Number of Hydrogens       & (integer)                                  \\\midrule
\multirow{2}{*}{Edge} & Bond type                 & Single, double, triple, aromatic (one-hot) \\
                      & Distance between atoms    & (real number)  \\\bottomrule                            
\end{tabular}
\end{threeparttable}
\small
\end{table}

\begin{figure}[tbp]
	\centering
    \includegraphics[width=1.\textwidth]{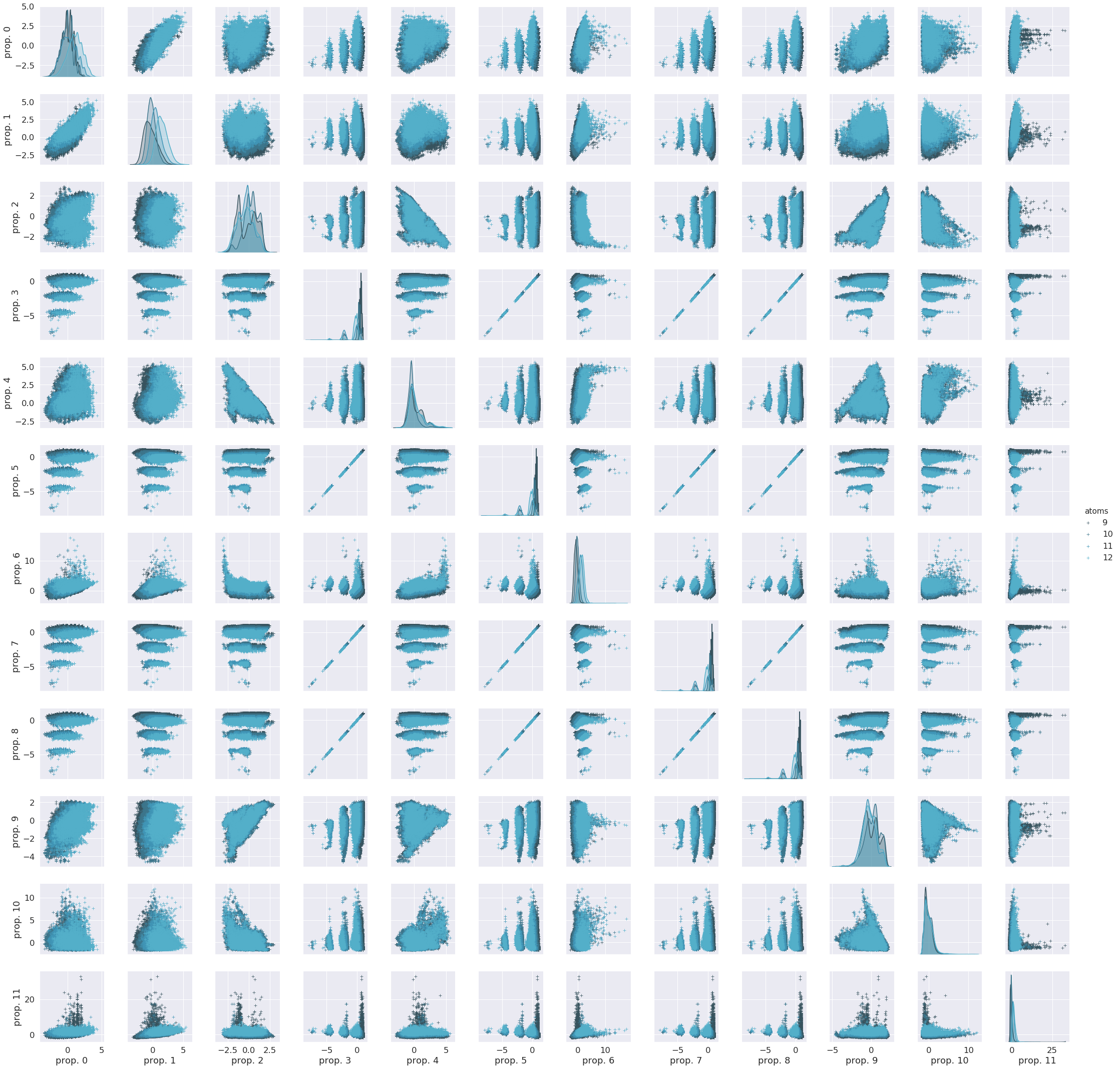}
	\small
	\caption{A pairplot for calculated properties.}
	\label{fig:prop}
\end{figure}

In both benchmarks, we use the Alchemy dataset which contains 119,487 molecules.  All node and edge features used in the benchmarking experiments are listed in Table~\ref{tab:feat}. For all models, we use as many features as compatible with a given model's setup: the input node feature is 15 dimensional for all models; As for the edge feature, GCN, ChebyNet, GAT and GIN only consider the binary feature of the existence of a bond. RGCN, GGNN, LanczosNet, and MPNN further incorporate bond types into the model. MPNN* takes input of both categorical bond type and continuous distance as edge features. For LanczosNet and GGNN, the implementation of the readout function is a simple attention based pooling which follows exactly the prescription given in the original papers. For all other models, we use the same readout, set2set~\citep{vinyals2015order}, which has more learnable parameters and is shown to work well~\citep{gilmer2017neural} in practice.  We jointly train 12 target tasks~(each task corresponds to predicting one of the molecular properties in Table~\ref{tab:dft}) for all models.
The targets are normalized to zero mean and unit variance with respect to this subset of 119,487 molecules. 
As we will host a molecular property prediction challenge, those 12 properties are randomly shuffled and anonymized in Tables~\ref{tab:exp_stratified} and \ref{tab:tab_random}. The pairplot of calculated properties are shown in Figure~\ref{fig:prop}, where the diagonal elements correspond to the distribution of each property and the off-diagonal elements correspond to the pairwise joint-distribution.

\begin{table}[t]
\centering
\scriptsize
\begin{threeparttable}
\caption{Performance Comparison on Alchemy~(Stratified Split) by the average MAE of all tasks and the separate MAE of each task~(Each task is to predict one of the properties in Table~\ref{tab:dft}).}\label{tab:exp_stratified}
\begin{tabular}{ccccc|cccc|c}
\toprule
\multirow{2}{*}{} & \multicolumn{4}{c|}{\multirow{2}{*}{Binary edge}} & \multicolumn{4}{c|}{\multirow{2}{*}{Bond type}} & \multirow{2}{*}{\begin{tabular}[c|]{@{}l@{}}Bond type \\\& distance\end{tabular}} \\
                  & \multicolumn{4}{c|}{}                        & \multicolumn{4}{c|}{}                             &                                                                               \\
         \cmidrule{2-10}
        & GCN    & {ChebyNet} & {GAT} & {GIN}   & {RGCN}   & {GGNN}   & {LanczosNet} & {MPNN}   & {MPNN*}\\
        \cmidrule{2-10}
MAE     & 0.1602 & 0.1636   & 0.1566 & 0.1417 & 0.1403 & 0.1421 & 0.1557     & 0.1355 & 0.0451 \\
\midrule
task 0  & 0.0841 & 0.1028   & 0.0785 & 0.0721 & 0.0617 & 0.0457 & 0.0769     & 0.0549 & 0.0210 \\
task 1  & 0.1665 & 0.1782   & 0.1550 & 0.1385 & 0.1320 & 0.1339 & 0.1593     & 0.1228 & 0.0459 \\
task 2  & 0.1830 & 0.1800   & 0.1804 & 0.1657 & 0.1665 & 0.1773 & 0.1790     & 0.1642 & 0.0733 \\
task 3  & 0.0431 & 0.0514   & 0.0424 & 0.0407 & 0.0322 & 0.0266 & 0.0503     & 0.0314 & 0.0163 \\
task 4  & 0.2721 & 0.2694   & 0.2670 & 0.2444 & 0.2493 & 0.2623 & 0.2532     & 0.2464 & 0.0928 \\
task 5  & 0.0431 & 0.0514   & 0.0424 & 0.0407 & 0.0322 & 0.0266 & 0.0503     & 0.0314 & 0.0163 \\
task 6  & 0.1581 & 0.1687   & 0.1541 & 0.1392 & 0.1330 & 0.1315 & 0.1560     & 0.1272 & 0.0564 \\
task 7  & 0.0431 & 0.0514   & 0.0424 & 0.0407 & 0.0322 & 0.0266 & 0.0503     & 0.0314 & 0.0163 \\
task 8  & 0.0431 & 0.0514   & 0.0424 & 0.0407 & 0.0322 & 0.0266 & 0.0503     & 0.0314 & 0.0163 \\
task 9  & 0.1979 & 0.1923   & 0.1905 & 0.1748 & 0.1805 & 0.1888 & 0.1882     & 0.1761 & 0.0776 \\
task 10 & 0.4020 & 0.3953   & 0.3971 & 0.3688 & 0.3813 & 0.3943 & 0.4018     & 0.3750 & 0.0819 \\
task 11 & 0.2859 & 0.2705   & 0.2870 & 0.2337 & 0.2501 & 0.2651 & 0.2531     & 0.2343 & 0.0277 \\
\bottomrule
\end{tabular}
\end{threeparttable}
\end{table} 

\begin{table}[t]
\centering
\scriptsize
\begin{threeparttable}
\caption{Performance Comparison on Alchemy~(Size Split)  by the average MAE of all tasks and the separate MAE of each task~(Each task is to predict one of the properties in Table~\ref{tab:dft}).}\label{tab:tab_random}
\begin{tabular}{ccccc|cccc|c}
\toprule
\multirow{2}{*}{} & \multicolumn{4}{c|}{\multirow{2}{*}{Binary edge}} & \multicolumn{4}{c|}{\multirow{2}{*}{Bond type}} & \multirow{2}{*}{\begin{tabular}[c|]{@{}l@{}}Bond type \\\& distance\end{tabular}} \\
                  & \multicolumn{4}{c|}{}                        & \multicolumn{4}{c|}{}                             &                                                                               \\
         \cmidrule{2-10}
        & GCN    & {ChebyNet} & {GAT} & {GIN}    & {RGCN}   & {GGNN}   & {LanczosNet} & {MPNN}   & {MPNN*}\\
        \cmidrule{2-10}
MAE     & 0.2542 & 0.2598   & 0.2304 & 0.2157 & 0.1843 & 0.1862 & 0.2233     & 0.1808 & 0.0655 \\
\midrule
task 0  & 0.1829 & 0.1989   & 0.1443 & 0.1382 & 0.0795 & 0.0682 & 0.1396     & 0.0692 & 0.0303 \\
task 1  & 0.2976 & 0.3209   & 0.2533 & 0.2369 & 0.1532 & 0.1605 & 0.2421     & 0.1513 & 0.0647 \\
task 2  & 0.2229 & 0.2170   & 0.2164 & 0.1998 & 0.1948 & 0.2014 & 0.2102     & 0.1968 & 0.0940 \\
task 3  & 0.1129 & 0.1244   & 0.0906 & 0.0890 & 0.0555 & 0.0464 & 0.0970     & 0.0506 & 0.0272 \\
task 4  & 0.3499 & 0.3416   & 0.3346 & 0.3191 & 0.3164 & 0.3209 & 0.3252     & 0.3212 & 0.1220 \\
task 5  & 0.1129 & 0.1244   & 0.0906 & 0.0890 & 0.0555 & 0.0463 & 0.0976     & 0.0506 & 0.0272 \\
task 6  & 0.3310 & 0.3488   & 0.2785 & 0.2722 & 0.2053 & 0.2043 & 0.2843     & 0.2055 & 0.0938 \\
task 7  & 0.1129 & 0.1244   & 0.0906 & 0.0890 & 0.0555 & 0.0464 & 0.0967     & 0.0506 & 0.0272 \\
task 8  & 0.1129 & 0.1244   & 0.0906 & 0.0890 & 0.0555 & 0.0463 & 0.0963     & 0.0506 & 0.0272 \\
task 9  & 0.2401 & 0.2409   & 0.2385 & 0.2162 & 0.2196 & 0.2345 & 0.2327     & 0.2191 & 0.1023 \\
task 10 & 0.5129 & 0.5005   & 0.4978 & 0.4741 & 0.4775 & 0.4917 & 0.4911     & 0.4723 & 0.1282 \\
task 11 & 0.4608 & 0.4514   & 0.4391 & 0.3764 & 0.3433 & 0.3672 & 0.3668     & 0.3314 & 0.0425 \\
\bottomrule
\end{tabular}
\end{threeparttable}
\end{table}

\subsection{Stratified Split}
Stratified split ensures each of the training, validation, and test sets to cover the full range of provided labels. Among properties in Table~\ref{tab:dft}, HOMO-LUMO gap is a particularly insightful quantity for understanding the photochemistry, photophysics, and single electron transfer mechanism of organic molecules. Because of its importance, we choose the HOMO-LUMO gap as the criterion to perform the stratified split.

We first sort samples in order of ascending value of HOMO-LUMO gap, then the ordered data is split into contiguous subsets with 10 samples each. For each subset, we randomly assign samples into the training, validation and test sets with the 8/1/1 ratio. Finally, we have 95591, 11948, 11948 molecules, respectively, in these 3 sets. Results of stratified split are reported in Table~\ref{tab:exp_stratified}. MPNN* stands out since it can utilize the information of atomic pairwise distances. As for other models, it is unclear how to concurrently use both continuous (atomic pairwise distances) and discrete~(chemical bond type) edge attributes. Without distance information, all models perform comparably. We note that the best result of each task is achieved by different models.

\subsection{Size Split}
We next consider a split of data based on the molecular size: the number of heavy atoms in each molecule. According to an empirical observation that, over the years, there has been a steady rise in the molecular weight (another indicator of molecular size) of medicinal chemistry compounds. Therefore, it is also meaningful to investigate how well a model trained mostly with smaller-sized molecules perform when it is used to predict molecules of bigger size.

The size-split experiment is conducted in the following approach. We reserve a subset of molecules comprising of either 11 or 12 heavy atoms for validation and testing. Specifically, we reserve 3,951 molecules as the validation set and 15,760 molecules as the test set. Finally, the remaining 99,776 molecules constitute the training set. In the training set, there are only 5,840 molecules with more than 10 heavy atoms. Hence, all models are mostly trained with molecules comprising of either 9 or 10 heavy atoms.

Similar to the results reported in the stratified-split experiment, MPNN* stands out again among the pack while all other models perform comparably. 
Nevertheless, the difficulty of the size split is obvious. All model performance drop in comparison to the stratified split. This performance hit implies the generalizability of these ML models could be further improved.







\section{Conclusions and Discussions}
\label{sec:conclusion}

  The process of building the Alchemy dataset has been a rewarding experience. Along this journey, we face the difficulty to optimize molecular geometries efficiently and accurately.  We eventually use a combination of two geometry optimization tools to overcome this problem. All the discrepancies between our data and QM9's can be precisely attributed to the disagreements on the optimized geometries. 


Our two insightful benchmarking results clearly manifest the superiority of GNN models that could best use additional molecular features such as atomic pairwise distances. The size-split benchmarking experiment suggests one direction for future model improvements.
The public availability of this dataset shall generate additional insights and further improvements of ML models for molecular sciences. 



\section{Alchemy Contest}
To engage more researchers to use our dataset and facilitate the development of better ML models for molecular sciences, we are hosting a molecular property prediction challenge based on the Alchemy dataset, named the Alchemy contest \footnote{https://alchemy.tencent.com}. The setting of our contest follows the size split in Section 5.2. The main task is to predict $12$ quantum mechanism properties of around 20K molecules given around 100K training samples, which primarily comprises of molecules with $9$ and $10$ heavy atoms and few molecules with more than $10$ heavy atoms. The testing samples are molecules mainly composed of $11$ and $12$ heavy atoms. This challenge is designed to mimic some realistic scenarios in which scientific developments progress by discovering more and more complex (such as larger size) molecules useful for a specific task. Therefore, it is desirable to build models that could predict properties of larger-sized molecules based on learning the performance of smaller-sized ones. Many modern ML techniques can be adopted for this contest; for instances, the few-shot learning is to learn knowledge by accessing a few training molecules composed of $11$ and $12$ heavy atoms, and the transfer learning is to apply the knowledge learned from smaller-size molecules to study the larger-size ones etc.

Our contest consists of two phases. In the first phase (May 22, 2019 - July 31, 2019), we release a training set of around 100K molecules for model developments and around 4K molecules for validation. In the second phase (August 1, 2019 - September 1, 2019), we will release another around 16K molecules for the final evaluation. To make this a fair competition, we only allow one submission per day and a total of twenty submissions in the second phase. Last but not least, a cash prize (total \textyen$100,000$ RMB, around $\$ 15,000$ USD) will be awarded to the top three entries on the leaderboard in the Phase 2 only. At the end of 2019, we will publicly release the alchemy dataset, composed of 130,000+ molecules, useful for the development of ML models for molecular and material sciences.

\clearpage
\bibliography{alchemy_neurips2019}
\bibliographystyle{apalike}

\end{document}